%% file: main.tex
\documentclass[sigconf, review=false]{acmart}

\usepackage{booktabs} 
\usepackage{multirow}
\setcopyright{none}

\acmDOI{10.475/123_4}

\acmISBN{123-4567-24-567/08/06}

\acmConference[ICVGIP'24]{15th Indian Conference on Computer Vision, Graphics and Image Processing}{December 2024}{Bengaluru, India}
\acmYear{2024}
\copyrightyear{2024}

\acmPrice{15.00}

\begin{document}
\title{WavShadow: Wavelet Based Shadow Segmentation and Removal}








\author{
    Shreyans Jain$^\dagger$ \quad Viraj Vekaria$^\dagger$ \quad Karan Gandhi$^\dagger$ \quad Aadya Arora$^\dagger$ \\
    \textit{Indian Institute of Technology Gandhinagar, Gujarat, India} \\
}
\thanks{$\dagger$ All authors contributed equally to this work.}

\renewcommand{\shortauthors}{}

\begin{abstract}
Shadow removal and segmentation remain challenging tasks in computer vision, particularly in complex real-world scenarios. This study presents a novel approach that enhances the ShadowFormer model by incorporating Masked Autoencoder (MAE) priors and Fast Fourier Convolution (FFC) blocks, leading to significantly faster convergence and improved performance. We introduce key innovations: (1) integration of MAE priors trained on Places2 dataset for better context understanding, (2) adoption of Haar wavelet features for enhanced edge detection and multi-scale analysis, and (3) implementation of a modified SAM Adapter for robust shadow segmentation. Extensive experiments on the challenging DESOBA dataset demonstrate that our approach achieves state-of-the-art results, with notable improvements in both convergence speed and shadow removal quality. 
\end{abstract}
%

\ccsdesc[500]{Computing methodologies~Image segmentation}

\keywords{Image Segmentation, Wavelet, Masked Auto Encoding}

\maketitle

\input{samplebody-conf}

\bibliographystyle{ACM-Reference-Format}
\bibliography{ICVGIP-Latex-Template}

\end{document}

%% file: samplebody-conf.tex
    \section{Introduction}
\noindent
Shadow removal and segmentation are essential in computer vision applications like image enhancement and object detection. These tasks are challenging in complex real-world scenarios where shadows intricately interact with objects.
\noindent
In this paper, we present a novel approach that combines Masked Autoencoder (MAE) priors and  advanced segmentation techniques to improve shadow removal and segmentation. Our method builds upon the ShadowFormer model, integrating MAE priors to achieve faster convergence and enhanced performance. We evaluate our approach using the DESOBA dataset, which offers a more realistic challenge compared to simpler datasets by including both object and shadow information.
\noindent
A key innovation in our work is the integration of a state-of-the-art SAM (Segment Anything Model) adapter for shadow segmentation. This adapter generates masks that are used in the shadow removal process, offering better generalization than ground truth masks. Additionally, we incorporate 'haar' wavelet features  into our approach, further improving the accuracy of shadow segmentation.
\noindent
Our research demonstrates the potential of combining advanced priors and segmentation techniques to significantly enhance the effectiveness of shadow removal and segmentation in complex, real-world environments. The findings presented here open new avenues for improving various computer vision tasks that rely on accurate shadow handling.

\section{Related Works}

\begin{figure}[!t]
    \centering
    \includegraphics[width=0.8\linewidth]{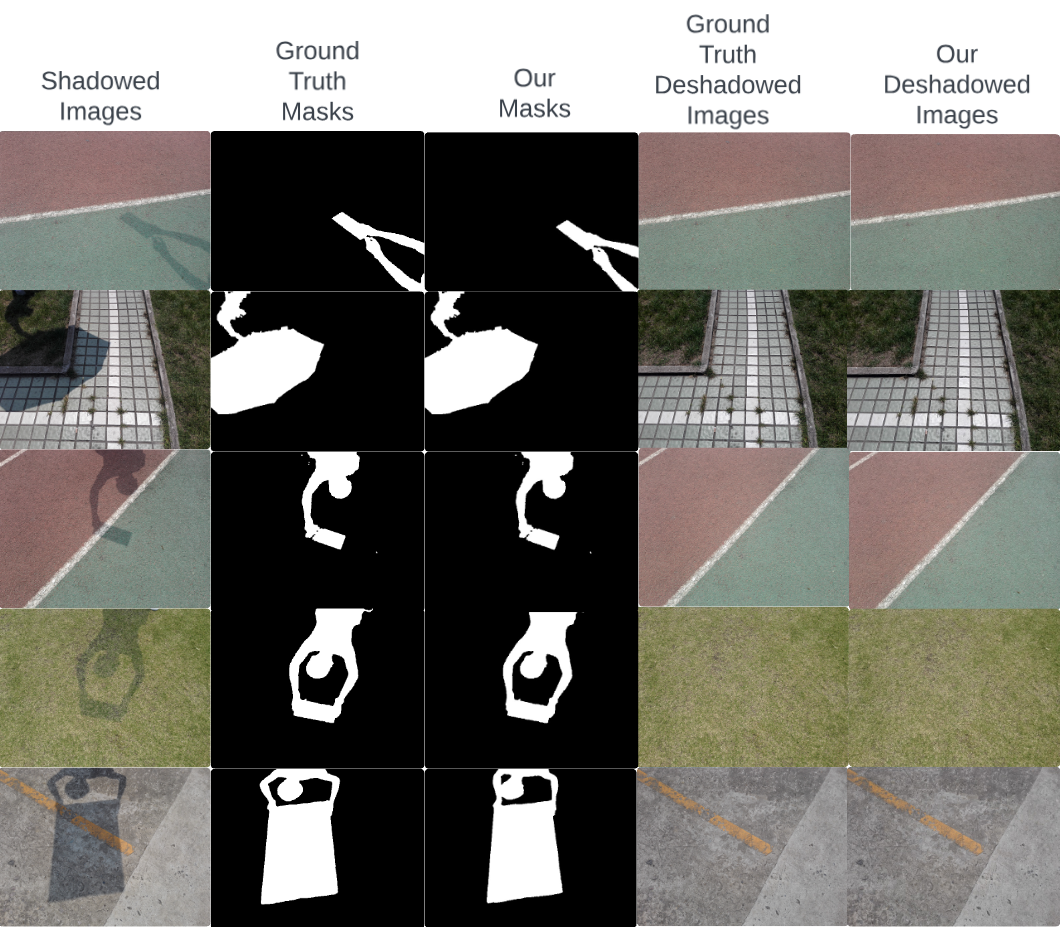}
    \caption{Qualitative Results. The figure shows shadow removal results using the proposed approach.}
    \label{fig:enter-label}
    \vspace{-2mm}
\end{figure}

\textbf{Traditional Shadow Detection.}  Early methods used physical models and heuristics based on chromacity, intensity, and texture, such as Salvador et al.'s spectral and geometrical properties ~\cite{salvador2004cast} and Tian et al.'s spectral power distribution differences ~\cite{tian2016new}.

\begin{table*}[!t]
\centering
\caption{Comparison of various methods across different regions}
\vspace{-3mm}
\resizebox{\linewidth}{!}{%
\begin{tabular}{|l|l|c|c|c|c|c|c|c|c|c|}
\hline
& & \multicolumn{3}{c|}{Shadow Region} & \multicolumn{3}{c|}{Non-Shadow Region} & \multicolumn{3}{c|}{All} \\
\cline{3-11}
& Dataset& PSNR & SSIM & RMSE & PSNR & SSIM & RMSE & PSNR & SSIM & RMSE \\
\hline
\multirow{2}{*}{ShadowFormer(L)}& ISTD & $\textbf{37.065}$ & $0.985$ & $4.815$ & $32.244$ & $0.952$ & $\textbf{4.418}$& $\textbf{30.581}$& $\textbf{0.934}$& $4.815$ \\
\cline{2-11}
& DESOBA & $37.456$& $0.982$& $13.004$& $44.788$& $\textbf{0.997}$& $0.311$& $36.291$& $\textbf{0.972}$& $0.963$\\
\hline
\multirow{2}{*}{MAE+FFC+SF} & ISTD & $36.777$& $\textbf{0.986}$& $7.598$& $32.078$& $0.951$& $4.446$& $30.328$& $0.932$& $4.897$\\
\cline{2-11}
& DESOBA & $\textbf{37.961}$ & $\textbf{0.983}$ & $\textbf{10.904}$ & $\textbf{44.943}$ & $\textbf{0.997}$ & $\textbf{0.208}$ & $\textbf{36.671}$ & $\textbf{0.972}$ & $\textbf{0.753}$ \\
\hline
\multirow{2}{*}{MAE+SF} & ISTD & $32.222$ & $0.951$ & $\textbf{4.361}$& $\textbf{36.694}$& $\textbf{0.985}$ & $7.649$ & $30.450$ & $0.932$ & $\textbf{4.814}$\\
\cline{2-11}
& DESOBA & $37.135$ & $0.982$ & $13.550$ & $44.665$ & $\textbf{0.997}$& $\textbf{0.208}$& $36.670$ & $0.971$ & $\textbf{0.886}$\\
\hline
\end{tabular}%
}
\label{tab:comparison}
\vspace{-0.5mm}
\end{table*}

\noindent
\textbf{Machine Learning Approaches.} Machine learning advanced shadow detection by using manually crafted features (texture, color, edge) with classifiers like SVM and decision trees. While innovative, these methods were limited in fully characterizing shadows in general scenarios ~\cite{guo2011single,vicente2015leave,vicente2018leave}.

\noindent
\textbf{Deep Learning Approaches.} Deep learning transformed shadow detection through automatic feature extraction. Key developments include Khan et al.'s pioneering use of CNNs ~\cite{khan2014automatic}, Hu et al.'s direction-aware spatial context learning ~\cite{DBLP:journals/pami/HuFZQH20,Hu_2018_CVPR}, Wang et al.'s iterative shadow detection and removal using GANs ~\cite{wang2018stacked}, Recent advancements in bidirectional feature pyramid networks ~\cite{zhu2018bidirectional} and distraction-aware shadow detection ~\cite{zheng2019distraction}.

\noindent
\textbf{Segment Anything Model and Adaptations.} The Segment Anything Model (SAM) ~\cite{kirillov2023segment} brought versatility to image segmentation. Chen et al.'s SAM-Adapter ~\cite{segmentanything} enhanced its performance in shadow detection by incorporating domain-specific information.

\noindent
\textbf{Instance Shadow Detection and Removal.} Instance shadow detection aims to identify individual shadows and their corresponding objects. Recent developments include Deep learning techniques for shadow elimination in various contexts ~\cite{qu2017deshadownet,hu2019mask,ding2019argan}, CondInst architecture for instance segmentation ~\cite{tian2020conditional}, Wang et al.'s bidirectional relation learning for shadow-object interactions ~\cite{Wang_2022}.

\vspace{-3 mm}
\section{Methodology}
Our proposed methodology integrates advanced feature extraction techniques with state-of-the-art models to enhance shadow removal and segmentation performance. The approach comprises the following key components:

\noindent
\textbf{Feature Extraction with Wavelet Transforms.} To effectively capture the intricate patterns and high-frequency details associated with shadows, we employ the Haar wavelet transform on the input images. We chose the Haar wavelet transform for its computational efficiency, ability to capture edge information at shadow boundaries, and multi-resolution analysis capability that helps distinguish between soft and hard shadow regions. By decomposing images into multiple frequency components, the Haar transform enables extraction of both spatial and frequency information while preserving important discontinuities in shadow regions. By integrating these Haar wavelet features into the Segment Anything Model (SAM) Adapter network, we enhance the network's ability to discern subtle shadow nuances, thereby improving mask generation accuracy.

\noindent
\textbf{Adapter Network Modification.}  We apply the Haar wavelet transform to input images by decomposing images into multiple frequency components, enabling the extraction of both spatial and frequency information. Integrating these wavelet features into the SAM Adapter enhances its ability to generate accurate shadow masks by leveraging multi-scale information.

\noindent
\textbf{Shadow Removal Pipeline Using Masked Autoencoder (MAE).} For the shadow removal task, we utilize a pre-trained Masked Autoencoder (MAE) trained on the Places2 dataset. The pipeline operates as follows:

\begin{enumerate}
    \item \textbf{Input Processing:} The shadowed image \( I \) and the corresponding shadow mask \( M \) generated by the modified SAM Adapter are fed into the MAE.
    \item \textbf{Prior Generation:} The MAE processes these inputs to produce an intermediate output \( I' \), which serves as a contextual prior. This prior encapsulates essential information about the scene, aiding in the reconstruction of a shadow-free image. We simply concatenate the Image priors in the input.
    \item \textbf{Integration with ShadowFormer:} The prior \( I' \) and the shadow mask \( M \) are then provided to the ShadowFormer model. ShadowFormer leverages these inputs to generate the final shadow-free image \( I_{\text{free}} \), ensuring both the removal of shadows and the preservation of image fidelity.
    \item \textbf{Using FFC Blocks to better encapsulate Global Context:} We used Fast Fourier Convolution blocks introduced in the LaMa Model \cite{suvorov2021resolution} in the Shadow Interaction Module of the ShadowFormer to better encapsulate Global Context.
\end{enumerate}

\noindent
\textbf{Training and Optimization.} The Sam Adapter and the Shadowformer with MAE prior are trained separately using the DESOBA and ISTD datasets, which offers complex real-world scenarios with intricate shadow-object interactions. Key aspects of the training process include:

\begin{itemize}
    \item \textbf{Loss Functions and Optimizers:} We used the same losses and optimizers to train the MAE, ShadowFormer and the SAM-Adapter.
    \item \textbf{Evaluation Metrics:} Performance is assessed using  Peak Signal-to-Noise Ratio (PSNR), Structural Similarity Index Measure (SSIM), and Root Mean Square Error (RMSE).
\end{itemize}

\noindent
\textbf{Implementation Details.} The MAE is pre-trained on the Places2 dataset, providing robust feature representations that enhance the shadow removal process. The DESOBA dataset is selected for its realistic and challenging shadow scenarios, ensuring that the model generalizes well to real-world applications. We have trained the network on NVIDIA V100 GPUs.
\vspace{-3 mm}
\section{Results}

\noindent
Integrating wavelet features with the SAM-adapter resulted in faster convergence and improved performance. Similar enhancements were observed on more complex datasets, such as DESOBA. Likewise, when incorporating MAE priors and FFC features into ShadowFormer, the model converged in approximately $250$ epochs, significantly earlier than the original model, which required around $450$ epochs. Our modified version consistently achieved comparable, and in some cases superior, performance compared to the vanilla ShadowFormer.
\vspace{-3 mm}

\section{Conclusion}
Our approach of integrating MAE priors, SAM-Adapter, and wavelet features significantly improves shadow removal and segmentation performance, particularly in complex real-world scenarios. The proposed method demonstrates faster convergence and enhanced accuracy, opening new possibilities for advanced shadow handling in computer vision tasks.
